\documentclass[AMS,STIX1COL]{WileyNJD-v2}
\usepackage{amsmath,amsfonts}
\usepackage{caption}
\usepackage{booktabs}
\usepackage{varwidth}
\usepackage{natbib}

\articletype{Article}%

\received{-}
\revised{-}
\accepted{-}

\begin{document}

\title{Feature Selection for Imbalanced Data with Deep Sparse Autoencoders Ensemble}

\author[1,2]{Michela C. Massi}
\author[1,2,3]{Francesca Ieva}
\author[4]{Francesca Gasperoni}
\author[1,2]{Anna Maria Paganoni}

\address[1]{MOX Laboratory for Modeling and Scientific Computing,  Department of Mathematics, Politecnico di Milano}
\address[2]{CADS - Center for Analysis Decisions and Society, Human Technopole}
\address[3]{CHRP - Center for Healthcare Research and Pharmacoepidemiology, Bicocca University}
\address[4]{MRC-Biostatistics Unit, University of Cambridge}

\authormark{Massi M.C. \textsc{et al}}

\corres{*Michela Carlotta Massi, Department of Mathematics, Politecnico di Milano, Via Edoardo Bonardi 9, 20133 Milano, Italy. \email{michelacarlotta.massi@polimi.it}}


\abstract[Summary]{Class imbalance is a common issue in many domain applications of learning algorithms. Oftentimes, in the same domains it is much more relevant to correctly classify and profile minority class observations. This need can be addressed by Feature Selection (FS), that offers several further advantages, s.a. decreasing computational costs, aiding inference and interpretability. However, traditional FS techniques may become sub-optimal in the presence of strongly imbalanced data. To achieve FS advantages in this setting, we propose a filtering FS algorithm ranking feature importance on the basis of the Reconstruction Error of a Deep Sparse AutoEncoders Ensemble (DSAEE). We use each DSAE trained only on majority class to reconstruct both classes. From the analysis of the aggregated Reconstruction Error, we determine the features where the minority class presents a different distribution of values w.r.t. the overrepresented one, thus identifying the most relevant features to discriminate between the two. We empirically demonstrate the efficacy of our algorithm in several experiments on high-dimensional datasets of varying sample size, showcasing its capability to select relevant and generalizable features to profile and classify minority class, outperforming other benchmark FS methods. We also briefly present a real application in radiogenomics, where the methodology was applied successfully.}



\maketitle

\footnotetext{\textbf{Abbreviations:} DSAEE, deep sparse autoencoder ensemble; FS, feature selection; AE, autoencoder; DSAE, deep sparse autoencoder; RE, reconstruction error; ML, machine learning; FSDS, feature selection dataset; CDS, classification dataset; LR, logistic regression; DT, decision tree; SVM, support vector machine; NB, naive bayes; NN nearest neighbor; AUROC, area under the roc curve; RFE, random feature elimination; LT, late toxicity}

\section{Introduction}
\label{intro}

A well-known problem of many real life applications of statistical models and machine learning algorithms is class imbalance \cite{anwar2014measurement}. Examples can be found in many sensitive domains such as medicine \cite{mazurowski2008training}, especially in case of rare disease classification tasks \cite{Hecoselection}, fraud detection \cite{wei2013effective}, fault detection \cite{zhu2010fault}, cyber security \cite{yousefi2017autoencoder} and many others \cite{ali2015classification}. All these domains share the same peculiarity: the importance of correctly identifying and profiling the minority class. In these contexts, a false negative is usually much more expensive w.r.t. a false positive. A straightforward example comes from the medical field, where a missed diagnosis in many cases is extremely risky for the patient's health and costly for the healthcare system \cite{annemans2013current, thabtah2020data}. \\
Moreover, on top of precise classification of minority class observations, domain experts are oftentimes interested in understanding which specific features (i.e. characteristics of their patients, or customers, etc.) should be kept under control or to investigate to drive decisions or invest in future research. The importance of identyfying the discriminant characteristics of the minority class is particularly evident in the clinical field, where an inaccurate feature selection can lead to an inaccurate diagnosis \cite{jung2020machine}. This observation holds for Genome Wide Association Studies for precision medicine \cite{hira2015review}, where the clinical interest lies on detecting the traits that are associated to a specific disease \cite{austin2013penalized}. Answering to this question, rather than merely classifying observations, gets harder as the number of features and the non-linearity of their interrelationships rises, driving growth in models' complexity.
One way of addressing this need is through Feature Selection (FS) techniques.
\\\\
In general, FS helps in identifying highly influential features that provide intrinsic information and discriminant property for class separability, while decreasing computational costs, aiding inference and giving better understanding on model representation \cite{guyon2003introduction,wasikowski2009combating}. However, it has been argued that traditional FS techniques become sub-optimal or even prejudicial to classification effectiveness when the classes are strongly imbalanced \cite{zheng2004feature, yin2013feature}. In \cite{yin2013feature}, the authors demonstrate through a simulation study how the overlapping of the classes' distributions after feature selection increases because of the strong bias towards the majority class, hindering classification performance. Therefore, to achieve the advantages granted by FS, a method tailored to address imbalanced settings without affecting classification accuracy is desirable.\\
Indeed, we argue that a FS robust to class imbalance can address both needs for accurate classification of the underrepresented class and for the identification of the specific pieces of information that are the most relevant for its identification. In other words, by selecting the most informative features to discriminate between classes, such a FS method can serve as a useful tool for the task of \textit{minority class profiling}. In Section \ref{radiogenomics} we will briefly describe a real case study where the methodology presented in this work successfully played this role in a complex research setting.
Nonetheless, although FS for imbalanced classification is recently gaining momentum, the number of reported works on the subject is still limited \cite{ali2015classification}. Few contributions dealt with this multi-faceted problem \cite{Hecoselection}.\\\\
For these reasons, in this work we focused on developing a novel FS method tailored to identify relevant features to discriminate the minority from the majority class in strongly imbalanced binary classification settings. In order to accomplish this task, in this paper we propose a filtering algorithm that ranks feature importance on the basis of a Deep Sparse AutoEncoders Ensemble (DSAEE). \\\\
From a methodological standpoint, the value provided by our proposal comes from the combination of two aspects: on the one hand, the choice of a particular type of AutoEncoder (AE) \cite{hinton2006reducing} as underlying model, on the other, the inclusion of this model within an ensemble algorithm.\\
Indeed, AEs are Neural Network (NN) models capable of flexibly capturing non-linear relationships among features \cite{han2018autoencoder}. These models have been exploited as feature selectors but, to the best of our knowledge, never tailored to class imbalance (cfr. Section \ref{sec:aefs}). Here we claim they can be effectively exploited as feature selectors specifically for an imbalanced setting if we consider the duality between imbalanced minority class classification and outlier detection. Indeed, as the minority class is rare w.r.t. the majority one, its observations might be considered outliers w.r.t. the normal population (inliers) constituted by the overrepresented class. AEs were previously recognized as powerful reconstruction-based outlier detection methods \cite{ma2013parallel, protopapadakis2017stacked, chen2017outlier, chen2018evolutionary, kieu2019outlier, sarvari2019unsupervised} that rely on scoring outliers by aggregating the Reconstruction Error (RE) for each observation. In this work, we propose to repurpose this reconstruction-based outlier detection approach to solve the problem of feature selection in imbalanced setting instead. Indeed, we apply an AE trained only on majority class observations to reconstruct both majority and minority classes: from the aggregation of the REs for each feature within each class, we determine where the minority class has a different distribution of values w.r.t. the majority class - thus identifying the most relevant features to discriminate between the two classes.\\
However, there exist the risk that a single AE fails in capturing the correlations among features, expecially in high dimensional settings \cite{chen2018evolutionary}, and a natural variance in results that might depend on the data, the design of the model and the local search for parameters typical of many Machine Learning (ML) methods. By using an ensemble approach as the one proposed in this work, and taking a central estimator of the RE, like the mean or the median, this variance is reduced \cite{dietterich2000ensemble, chen2017outlier}. Nevertheless, in order to make ensemble learning methods work, the individual ensemble components must be adequately diverse \cite{chen2017outlier, sarvari2019unsupervised}.
This is achieved in our proposition by designing the algorithm s.t. each ensemble component can capture different aspects of the underliyng majority class distribution.
In particular, the novelty of our approach resides in fostering this diversity among components through (i) a sampling procedure tailored for imbalanced settings that builds different training and test sets to supply to each learner, and (ii) a sparsity constraint imposed on the models.\\\\
In light of the above, the contributions of this work are multiple. We enlarge the limited literature on FS tailored to deal with the daunting real-life issue of class imbalance. We do that presenting an algorithm that repurposes the power of AEs as outlier detectors for reconstruction-based minority class-specific feature selection, which is a novelty for AE-based feature selectors in general. Finally, we robustify the selection thanks to its ensemble approach, designed to foster diversity of components and accuracy on minority class. 
\\\\
The remainder of the paper is organized as follows. In Section \ref{rw} we discuss some related works, strenghtening our positioning w.r.t. other approaches; in Section \ref{sec:methods} we provide some background on DSAEs, then we describe and discuss the proposed DSAEE algorithm in detail. In Section \ref{experimental} we describe a series of experiments and proof of concepts developed on several datasets of varying sample size and dimensionality: firstly we empirically validate the good performance of the selected feature subset despite the dimensionality reduction (Section \ref{improv}), then, we compare our proposed methodology with other state-of-the-art and more traditional FS methods (Section \ref{benchmark}). Additionally, we display some visualizations of the selected features to demonstrate their meaningfulness in discriminating minority from majority class (Section \ref{interpretation}) and finally we briefly describe an application on real clinical data (Section \ref{radiogenomics}). In Section \ref{conclusion} we highlight some relevant considerations on the proposed approach, and conclude with some final remarks and possible extensions.

\section{Related Works}
As stated in the introduction, in this paper we aim at presenting a novel FS method tailored to tackle class imbalance. Indeed, the method is designed to select a subset of informative features to reduce the impact of the strong imbalance between minority and majority classes on the classification performance. To frame the position of our proposal from a methodological point of view, in this section we will first describe other works developing methods to this aim. Then, as we are exploiting AEs as building blocks of our ensemble method for FS, we will report on studies that utilized these models for this task, irrespectively of the classes' distribution.
\label{rw}

\subsection{Feature Selection for Imbalanced Data}
In general, there are three approaches to apply FS algorithms in classification: wrapper, embedded and filter methods \cite{zhang2017feature}. Wrapper methods \cite{liu2007computational} make the FS revolve around the optimization of the performance of a predetermined classifier: the feature subset that maximises the defined performance metric is selected. In an imbalanced setting, the choice of the optimization metric is crucial. Indeed, among the available examples in the literature, some exploited the area under the ROC curve as a metric to select the best mix of features \cite{chen2008fast}, others the F-measure \cite{ali2015classification, zhang2017feature, liu2019classification}, while in \cite{maldonado2014feature} they exploit, among others, a balanced loss function which takes the weighted average of false positives and false negatives.
Despite their optimal results in terms of classification accuracy, wrapper methods are generally computationally expensive, and there is no guarantee of reaching a global optimum. \\
Embedded methods \cite{lal2006embedded} overcome this issue by determining the feature subset autonomously during classifier learning, by including for instance a regularization term in the loss function \cite{nie2010efficient}. However, to the best of our knowledge, no embedded method has been designed specifically to tackle class imbalance. An hybrid embedded and wrapper approach is instead proposed in \cite{liu2017cost}. Nonetheless, all the aforementioned methods are strictly bounded to a specific classifier.\\
Filter methods \cite{sanchez2007filter} are pre-processing algorithms that measure the usefulness of the feature subset for classification by working on the original data without involving any classifier. They usually rank features' importance on the basis of suitable metrics, some specifically tailored for imbalanced classification problems \cite{yin2013feature, zheng2004feature, cuaya2011minority}. Our proposal belongs to this classifier-agnostic type of algorithms.

\subsection{AutoEncoder-based Feature Selection}
\label{sec:aefs}
We will now provide a brief overview of how AutoEncoders (AEs) were employed as feature selectors in the available literature.\\
As mentioned, AEs \cite{hinton2006reducing} are a particular class of NNs widely used for learning of data representations \cite{baldi2012autoencoders}, dimensionality reduction \cite{hinton2006reducing} and outlier/anomaly detection \cite{aggarwal2015outlier, ma2013parallel, protopapadakis2017stacked,chen2017outlier, chen2018evolutionary, kieu2019outlier, sarvari2019unsupervised}.
This powerful representation learning method has been recently exploited for reconstruction-based feature selection as well. For instance, in \cite{chandra2015exploring} AEs are exploited as an unsupervised feature selection method, masking input features and using the Reconstruction Error (RE) of masked input features to compute feature weights in a moving average manner. In \cite{han2018autoencoder} the authors combine AE regression and a weight penalization on the input layer: feature importance is then derived from the value of the weights associated to each feature. Another sparsity-based unsupervised approach can be found in \cite{feng2018graph} and \cite{yu2019unsupervised}. Finally, in the most recent work in \cite{concrete}, the authors propose the Concrete AutoEncoder Feature Selector (CAEFS), that exploits the Concrete distribution to differentiate through the reconstruction loss and selects input features to minimize it.\\
All these approaches share an unsupervised setting and have demonstrated their potential as feature selectors against other state of the art techniques. Nonetheless, they all train one AE model only, incurring in the risks discussed in Section \ref{intro}. Moreover, they all are FS methods designed for balanced classification. This balanced selection of features was argued potentially harmful in strongly imbalanced settings \cite{zheng2004feature, yin2013feature}. What distinguishes our DSAEE from the available examples of AE-based feature selectors, is the ensemble approach to the problem - where each of the AE is one of a set of weak learners - and the tailoring of each model's training procedure inspired by outlier detection methods, to approach specifically imbalanced datasets.

\section{DSAE Ensemble (DSAEE) For Minority Class Feature Selection}
\label{sec:methods}
In Section \ref{sec:DSAE} we provide some background on the DSAE components and we detail the regularization we impose on the models to foster the diversity among each component. In Section \ref{sec:prop_alg} we detail how the proposed algorithm encapsulates each component into a tailored training procedure to identify the most relevant features to discriminate minority class in imbalanced settings.

\subsection{Background: AutoEncoders and Deep Sparse AutoEncoders}
\label{sec:DSAE}
An AE \cite{hinton2006reducing} is a NN trained to attempt to copy its input to its output. Let the matrix $\textbf{X} \in \rm I\!R^{N\times J}$ be the input data, $\textbf{X} = \{\textbf{x}_1, ...,\textbf{x}_N\}$ set of $N$ training vectors $i$ ($i \in \{1,...,N\}$), characterized by $J$ features. The shallow version of an AE is constituted by an input layer with $J$ nodes, a hidden layer with $H$ (with $H$ usually smaller than $J$) nodes that describes a \textit{code} used to represent the input, and an output layer of size $J$.
The network can be seen as constituted by two parts: an encoder and a decoder. The encoder function $\textbf{h}_{i}=f(\textbf{Wx}_{i}+\textbf{b})$, encodes each input vector $\textbf{x}_i$ into an encoded version of itself of size $H$. Here $f$ is usually non-linear and is referred to as \textit{activation function}, $\textbf{W} \in \rm I\!R^{H\times J}$ is called \textit{weight matrix} and $\textbf{b}$ is a $H$-dimensional \textit{bias} vector. The decoder maps back the encoded vector to the $J$-dimensional space in most cases using a squashing non-linear function $\hat{\textbf{x}_{i}} = g(\mathbf{W'h_{i}+b'})$, with parameters $\mathbf{W'} \in \rm I\!R^{J\times H}$ and $\mathbf{b'} \in \rm I\!R^{J}$. The model is trained through gradient descent of the loss function $ L(\textbf{x},\hat{\textbf{x}})$; where $L$ is typically the Mean Squared Reconstruction Error (MSRE), i.e. the mean squared Euclidean distance between the input values and the reconstructed values for each observation. Each training observation $\textbf{x}_{i}$ is thus mapped to a corresponding $\textbf{h}_{i}$ which is then mapped to a reconstruction $\hat{\textbf{x}}_{i}$ s.t. $\hat{\textbf{x}}_{i}$ $\approx$ $\textbf{x}_{i}$.\\
\indent To expand the shallow network to a \textit{deep} version, the formulation is similar, with the output of one layer being the input of the following layer.\\
Usually, AEs are built with constraints that force them not only to replicate the input, but to learn effective representations of such input in the hidden layer. One way to obtain useful representations from the autoencoer is to introduce sparsity in the code layer (Sparse AutoEncoders - SAE) by imposing a regularization term in the loss function. 
In order to do that, the model includes a sparsity penalty $\Omega(\textbf{h})$ on the hidden layer (or the most internal layer in case of deep architectures) $\textbf{h}$ additionally to the reconstruction error:
\[ L_{i} = L(\textbf{x}_{i}, \hat{\textbf{x}}_{i}) + \Omega(\textbf{h}_{i}). \]
The regularization can take various forms. In a deep architecture (Deep Sparse AutoEncoder - DSAE), let us consider $\textbf{h}_{i}^{(l)}$ as the activation of the most internal hidden layer ($l$) for the $i$-th observation vector $\textbf{x}_{i}$, i.e. the value of the function $\textbf{h}_{i}^{(l)} = f^{(l)}(\textbf{W}^{(l)} \textbf{h}_{i}^{(l-1)}+ \textbf{b}^{(l)})$. One way of obtaining a sparse representation is to add a penalty term that penalizes the $L_1$ norm of the vector $\textbf{h}_{i}^{(l)}$ for each observation $i$, controlled by a parameter $\lambda$, i.e.
\begin{equation}
\label{eq:loss+pen}
     L_{i} = L(\textbf{x}_{i}, \hat{\textbf{x}}_{i})+\lambda |\textbf{h}_{i}^{(l)}|.
\end{equation}
The parameter $\lambda$ can be optimized through grid search or can be arbitrarily chosen in the design phase of the model.\\
This penalization term forces the model to \textit{activate} the minimum number of hidden nodes to reconstruct the input. Paired with the input sampling described below, it increases the diversity among each learner in the ensemble. Moreover, it reduces the need for tailored choices or expensive optimization to define the proper architecture. 
\subsection{The Ensemble Algorithm}
\begin{figure}[t]
\centerline{\includegraphics[width=\columnwidth]{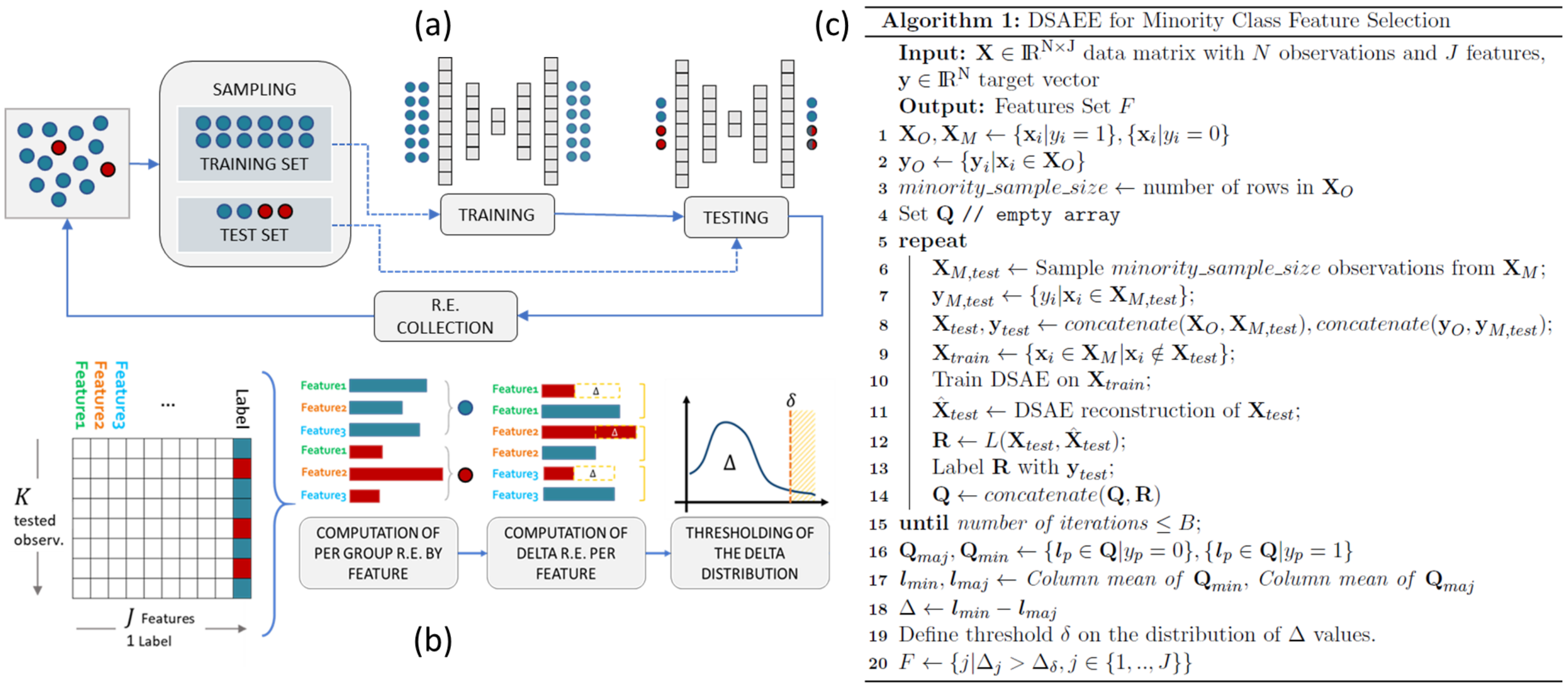}}
\caption{Training and FS of DSAEE algorithm. Panel (a) is a schema of the sampling procedure, repeated for each ensemble component. Panel (b) represents the steps of the algorithm from the concatenated $\textbf{Q}$ matrix of RE to the feature selection. Finally, panel (c) reports the pseudo-code of the whole DSAEE algorithm.}
\label{fig:flusso}
\end{figure}

\label{sec:prop_alg}
Let us consider the binary supervised learning setup with a training set of $N$ (input, target) pairs $D = \{(\textbf{x}_{1}, y_{1}),...,(\textbf{x}_{N}, y_{N})\}$ , where $y_{i}$ is the target that takes values is $\{0,1\}$ and $\textbf{X} \in \rm I\!R^{N\times J}$ is the input matrix. We consider the supervised learning to be imbalanced, thus the number of observations in the minority class ($O = \{\textbf{x}_{i}| y_{i}=1\}$) is relevantly smaller than the number of observations in the majority class ($M = \{\textbf{x}_{i}| y_{i}=0\}$). Our final objective consists in building a feature set F, with $|F|<J$ (from now on the notation $|\cdot|$ will represent the cardinality of a set), selecting the most relevant features to discriminate the minority from the majority class. We therefore define $\textbf{X}_{O} \in \rm I\!R^{|O| \times J}$ as the minority class observations and $\textbf{X}_{M} \in \rm I\!R^{|M| \times J}$ as the majority class ones.\\\\
With the intention of building an ensemble of different \textit{learners} from which to aggregate information to rank features, we first develop a tailored sampling procedure, inspired by the \textit{outlier detection} approaches, to train each learner on a different sample of data selected with the rationale detailed in the following, and schematized in Figure \ref{fig:flusso}(a).\\
In particular, from $\textbf{X}_{O}$ and $\textbf{X}_{M}$ and the respective outcomes $\textbf{y}_O$, and $\textbf{y}_M$ we generate a training set $\textbf{X}_{train}$ and a test set $\textbf{X}_{test}$. The test set contains $2|O|$ data points, including all the minority class observations, and an equal number of majority ones randomly drawn from $M$. The training set is instead composed by the majority class data excluded from the test set.\\ 
This structure of the two datasets allows us to train each DSAE learner in an unsupervised fashion only on the overrepresented population, and to test their performance when facing both majority and minority class examples, so that we can compare the RE made on the two populations. The rationale behind this sampling procedure is based on the fact that DSAEs trained to reconstruct \textit{normal} observations only (i.e. the majority class) will make higher RE when tested on \textit{outlier} observations (i.e. minority class examples) never experienced during training.\\
Indeed, once the two datasets are built, we train each DSAE on $\textbf{X}_{train}$ to minimize the loss function formulated in (\ref{eq:loss+pen}). Then, we supply $\textbf{X}_{test}$, we collect the reconstructed matrix $\hat{\textbf{X}}_{test}$, and for each $\textbf{x}_{p}$ in the test set, with $p \in \{1,..,P=2|O|\}$ we compute the vector of RE as the element-wise squared difference:
\begin{equation}
    \textbf{\textit{l}}_{p}(\textbf{x}_{p}, \hat{\textbf{x}}_{p}) = (\textbf{x}_{p}-\hat{\textbf{x}}_{p})^{2}.
\label{eq:vecRec_err}    
\end{equation}
We thus obtain a matrix of RE, $\textbf{R} = L(\textbf{X}_{test}, \hat{\textbf{X}}_{test})$, $\textbf{R} \in \rm I\!R^{P \times J}$  that has one row per observation $p$, and the $J$ features on the columns, that we label including $\textbf{y}_{test}$.\\
For $B$ ensemble learners included in the algorithm, we will produce $B$ sampled training and test sets, and concatenate the $B$ matrices $\textbf{R}$, building the final RE matrix $\textbf{Q} = \{(\textbf{\textit{l}}_{1}, y_{1}),...,(\textbf{\textit{l}}_{K}, y_{K}) \} \in \rm I\!R^{K \times (J+1)}$, where $K = PB$ is the total number of tested observations $p$ (now $p \in \{1,...,K\}$) and $(J+1)$ is the number of features plus the label associated to each observation. \\
As previously mentioned, we expect each AE to make higher average RE on the observations originally belonging to the group $O$ of minority class observations not evaluated during training. We can also consider each value $\textit{l}_{pj}$ in the vector $\textbf{\textit{l}}_{p}$, i.e. the RE committed on feature $j$ for the observation $p$. In this case, if the observation $p$ belongs to $O$, we would expect the model to make higher RE on the features where the minority class has a significantly different distribution of values w.r.t. the majority one. For a schema of the algorithm described in the following, refer to Figure \ref{fig:flusso}.(b).\\
In order to select the most representative features to discriminate between minority and majority class, we subdivide the vectors $\textbf{\textit{l}}_{p} \in \textbf{Q}$ in two matrices: one composed by minority class RE ($\textbf{Q}_{min}$) and the other by majority class RE ($\textbf{Q}_{maj}$).\\
From these sets we can estimate the vectors of average RE per feature $j$ per group: $\textbf{\textit{l}}_{min}$ and $\textbf{\textit{l}}_{maj}$, both belonging to $\rm I\!R^{J}$, where each element is computed as
\begin{equation}
    \textbf{\textit{l}}_{j,min}= \frac{1}{T}\sum_{t=1}^{T}\textbf{Q}_{tj,min},
    \label{eq:REbysnp1}
\end{equation}
\begin{equation}
    \textbf{\textit{l}}_{j,maj}= \frac{1}{T}\sum_{t=1}^{T}\textbf{Q}_{tj,maj},
    \label{eq:REbysnp2}
\end{equation}
and $T=K/2$ is the number of both minority and majority class examples in $\textbf{Q}$.
Once we have computed the class specific average REs per feature, we can proceed to the feature selection by studying how the RE of each feature varies between classes. To select only the features where the difference in RE is remarkable (i.e. where the minority class is notably distant from the majority one), we first compute the vector of $\Delta$ RE as
\begin{equation}
    \Delta = \textbf{\textit{l}}_{min} - \textbf{\textit{l}}_{maj}.
    \label{eq:deltadist}
\end{equation}
We observe the distribution of values taken by $\Delta$, to understand how distant minority class features are w.r.t. majority class ones. We can therefore define a quantile threshold $\delta$ on the $\Delta$ distribution. The observed $\Delta_{j}$ values above the defined quantile ($\Delta_{(\delta)}$) are considered relevant, and are therefore selected by the algorithm. In other words, we build the set of selected features $F$ including only the features $j$ with the highest difference in RE between the classes:
\begin{equation}
    F =  \{j| \Delta_j > \Delta_{\delta}, j \in \{1,..,J\}\} 
\end{equation}
From the original dataset \textbf{X} we can therefore extract a subset of features to either analyze per se or feed to any classifier. There is an inverse relation between $\delta$ and the number of selected features: the higher the $\delta$, the lower the numer of selected features.\\
Algorithm 1 in Figure \ref{fig:flusso}.(c) reports the pseudo-code of the whole FS procedure.

\subsection{Computational Complexity}
Each DSAEE component has a complexity $O(nwe)$  dependent on $n$ (the number of observations in the data matrix), $w$ (the number of weigths in the network) and $e$ (the number of \textit{epochs}, or iterations in the training). \\
The complexity of the training of an Ensemble of DSAEs becomes $\sim O(Bnwe)$, growing linearly with the number of $B$ trained models. Both the number of $B$ employed components and the architectural choices impacting $w$ and $e$ can be optimized to reduce training time and improve results as well. Moreover, the ensemble training can be easily parallelized, thus significantly cutting training time.

\section{Experiments}
\label{experimental}
To study the performance of the DSAEE and to test the validity of the claims raised in the previous sections we carried out several empirical evaluations.
In particular, we were interested in testing the capability of our algorithm to select even extremely small subsets of features while keeping the classification performance sufficiently high, especially on the minority class. This evaluation was carried out in settings of varying dimensionality and sample size (see Section \ref{improv}). Moreover, we compared the classification performance of our method against some benchmark FS algorithms (Section \ref{benchmark}) and finally, we investigated in an interpretable and visual way the meaningfulness of the selected features and their capability to provide useful insights to discriminate between minority and majority classes (Section \ref{interpretation}). 
To conclude, we also provide a brief description of a real data application in the challenging field of radiogenomics (see Section \ref{radiogenomics}). Through this analysis, we highlight the relevant impact that we are bringing in terms of minority class profiling in complex real-life research scenarios.

\subsection{Datasets and Performance Measures}
For all the aforementioned numerical experiments we decided to adopt freely distributed datasets to make results accessible and reproducible. Moreover, some peculiar characteristic of each of the exploited data allowed us to showcase different aspects of our algorithm and discuss its potential when applied to multifaceted scenarios. Note that the datasets exploited in our experiments were not originally imbalanced and in most cases they were meant for multiclass classification problems. As a consequence, a preliminary subsetting of the chosen data was conducted. In the following, we will list the adopted datasets and describe in details the dataset-building choices we made for each of them.\\
For all datasets, we selected one of the classes as the majority class, and we undersampled another class to represent the minority category. From the derived datasets, we extracted one subset on which we applied our feature selection method (\textit{Feature Selection} DataSet - FSDS), while the remaining was held out to evaluate the classification accuracy of the selected features (\textit{Classification} DataSet, CDS). In Table \ref{tab:datasets_proportions} we report all datasets, their composition, and the type of experiment they were exploited for.
\begin{enumerate}
    \item \textbf{ISOLET} \cite{fanty1991spoken} (number of observations $N = 370$; number of features $J = 617$). It consists of preprocessed speech data of people pronouncing the names of the letters in the English alphabet, and is widely used as a benchmark in the feature selection literature. Each feature is one of the 617 quantities produced as a result of the preprocessing. We chose class '\textit{A}' as the majority class, and '\textit{B}' as the minority one. Given the small number of observations available per class, this dataset allowed us to test the applicability of our algorithm in high dimensionality and small sample size settings.
    \item \textbf{GISETTE} \cite{guyon2005result} ($N = 3,300$; $J = 5,000$) This dataset was built for NIPS2003 feature selection challenge. The whole dataset contained 6,000 observations equally split between classes, with 5,000 features (50$\%$ of which are probes with no predictive power). We created 5 datasets including all 3,000 majority class observations and 300 randomly sampled minority class observations (9.05$\%$), and we splitted them into FSDS and CDS according to a 75/25 ratio.
    \item \textbf{Epileptic Seizure}  \cite{andrzejak2001indications} ($N = 11,500; 7,300$; $J = 178$). In this functional dataset, each data point represents 178 seconds of EEG recording for one of the 500 patients in the study. Each of the 178 features is the value of the EEG at that time-stamp. The label indicates whether the EEG is recording seizure activity ('Y') or not ('N'). This dataset was originally imbalanced, but we decided to increase the complexity by subsampling minority class further (cfr. Table \ref{tab:datasets_proportions}).
    \item \textbf{Fashion MNIST} \cite{xiao2017fmnist} ($N = [7,350; 7,300]$; $J = 784$). This dataset is composed by 28x28 grayscale images of clothing. To test our model we built two datasets with different imbalance rates. \textit{T-shirts} were selected as the majority class, and \textit{coats} as the minority one for the first dataset ($\sim 5\%$ of the whole dataset, with 7,350 total observations), while \textit{pullovers} for the second ($\sim 4\%$ with $N=7,300$). 
    \item \textbf{MNIST} \cite{MNIST} ($N = 8,292$; $J = 784$). This dataset is composed by 28x28 grayscale images of hand-written digits. We selected two quite overlapping classes to test our model: the '7' digit class as the minority class and the '1' digit class as the majority one. This dataset, together with the two extracted from Fashion MNIST, simulate a setting of extreme imbalance (below $95:5$ ratio) and moderately high dimensionality, but with a large sample size.

\end{enumerate}
It should be noted that the proposed algorithm is meant to be applied to features that do not present any dependence (i.e. the order of the features is irrelevant). Its applicability to image datasets is guaranteed by the fact that all images are centered, allowing us to meaningfully treat each pixel as an independent feature. The choice to add image datasets to these experiments derives from both their dimensionality and the clear readability of their results, that allow for visually investigating the selected features by representing them as pixels.\\
\indent To evaluate the classification performance in an imbalanced setting, we decided not to adopt the classical accuracy on both classes. Instead, we chose the Sensitivity metric (i.e.  the ratio of true positives and the sum of true positives and false negatives for observations belonging to the minority class) and the Area Under the Receiver Operating Characteristic (AUROC), that estimates the performance of a binary classifier comparing false positive rates with true positive rates and is a widely used metric to evaluate model's capability to correctly classify both classes, especially in imbalanced settings.

\begin{table}[t]
\begin{center}
\begin{small}
\begin{sc}
\resizebox{0.85\textwidth}{!}{%
\begin{tabular}{@{}lllll@{}}
\toprule
Dataset                        & Class        & FSDS - N (\%)         & CDS - N (\%)     & Experiments  \\ \midrule
ISOLET                        & 'a'         & 225 (81.9\%)   & 75 (80.7\%)  &   Feature Subset Performance    \\
\multicolumn{1}{c}{}          & 'b'          & 52 (18.1\%)    & 18 (19.3\%) & Benchmark  \\ \midrule
GISETTE                        & '0'         & 2,250 (90.91\%)   & 750 (90.91\%)      & Feature Subset Performance\\
\multicolumn{1}{c}{}          & '1'          &  225 (9.09\%)    & 75 (9.09\%) &  \\\midrule
Ep. Seizure                   & 'N'         & 6,440 (94.96\%)   & 2,760 (94.96\%)      & Feature Subset Performance\\
\multicolumn{1}{c}{}          & 'Y'          &  350 (5.04\%)    & 150 (5.04\%) &  \\\midrule
F-MNIST                       & T-shirts   & 5,250 (95.90\%)    & 1,750 (95.90\%)  & Feature Subset Performance  \\         
\multicolumn{1}{c}{}          & Pullovers   & 225 (4.10\%)    & 75 (4.10\%)   & Benchmark  \\ \cmidrule(l){2-5}
\multicolumn{1}{c}{}          & T-shirts     & 6,000 (95.3\%) & 1,000 (92.2\%) & Interpretability \\
\multicolumn{1}{c}{}          & Coats       & 300 (4.7\%)    & 50 (7.8\%)    \\\midrule
MNIST                          & '1'         & 6,742 (93.1\%) & 1,000 (92.2\%) & Interpretability \\
\multicolumn{1}{c}{}           & '7'          & 500 (6.9\%)    & 50 (7.8\%)     \\ 
\bottomrule
\end{tabular}
}%
\end{sc}
\end{small}
\end{center}
\caption{Feature Selection Dataset (FSDS) and Classification Dataset (CDS) composition for the datasets adopted in the experiments.}
\label{tab:datasets_proportions}
\end{table}

\subsection{Classification Performance of Selected Feature Subsets}
\label{improv}
Dimensionality reduction has impacts on computational time and complexity, noise reduction, model significance and results interpretability, but all these improvements should not come at the cost of a good classification performance on the classes of interest. In particular, in research scenarios as those presented in Section \ref{intro}, a minimum level of precision on Minority Class observations is desirable.\\
To test our algorithm, we applied it to the FSDS for various $\delta$ values ($\delta \in$ \{0.65, 0.7, 0.75, 0.8, 0.85, 0.9, 0.95, 0.97, 0.99\}), selecting different subsets of variables. For each $\delta$ we created from the CDS a dataset containing the selected features only. The CDS was subsequently subdivided in training set and test set according to a 70-30 split that was held constant across all experiments.\\
On the obtained datasets we trained and tested five classifiers: Logistic Regression (LR), Decision Tree (DT), Support Vector Machines (SVM), Naive Bayes (NB) and Nearest Neighbor (NN) classifier.  We chose to test different classifiers to verify whether our model-agnostic feature selection approach provided good results indepently of the subsequent classifier adopted. All algorithms were drawn from scikit-learn library for Python \cite{scikit-learn} and their hyperparameters were kept in default mode, unless differently stated. Note that we applied the same classifiers to all experiments without tailoring their parameters to the data at hand. This choice does not resemble a traditional classification process in a real-life scenario, where classifiers are optimized to improve the performance on the data at hand, but aimed at showcasing the impact of the feature subset selection alone.\\
Details on the code, the implementation and the specific architectural choices for the DSAEE are described and discussed in Appendix.\\
We tested the DSAEE feature selector on Isolet, Fashion-MNIST, Gisette and Epileptic Seizure datasets. 
\begin{figure}[t]
\centerline{\includegraphics[width=\columnwidth]{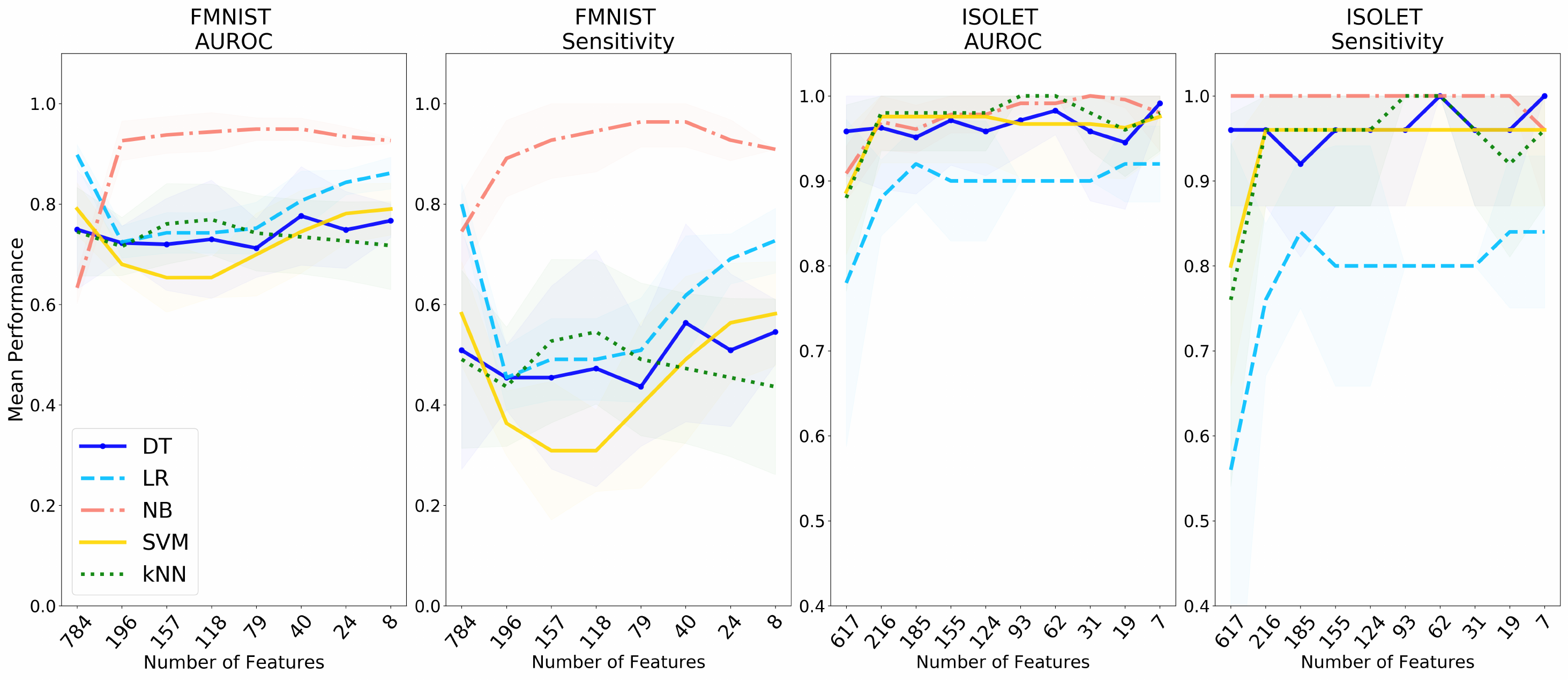}}
\caption{AUROC and Sensitivity improvement in the classification of FMNIST dataset (first two plots) and ISOLET dataset (second couple of plots) with different variables' subsets. The lines represent the average performance on 5 trials. Standard errors define the lighter areas around the line. The first number of features in each plot is the performance with the whole original features set.}
\label{fig:isolet_improv}
\end{figure}
\noindent Results for FMNIST and Isolet datasets are reported in Figure \ref{fig:isolet_improv}, where performance metrics are averaged over 5 trials and the x-axes display the size of the selected feature sets. \\
On FMNIST Dataset (Figure\ref{fig:isolet_improv}, first two panels) most classifiers suffered the dimensionality reduction up until smaller subsets, when their performance started growing again. On the contrary, NB classifier had a steep improvement on both AUROC and Sensitivity, reaching almost perfect scores for subsets of extremely small dimensionality (8 features, $\sim 1\%$ of the original 28x28 image).\\
On Isolet Data, where the sample size is extremely small compared to the number of features and the minority class in the training set contains 52 observations only, the five classifiers performed most of the times as good as the baseline performance with all variables, despite the reducing size of the features subset. Sensitivity (Figure \ref{fig:isolet_improv} fourth panel) increased substantially for KNN and SVM, while the LR classifier kept attaining an almost perfect score even as the cardinality ($|F|$) of the selected features set decreased substantially (while improving on AUROC score, as shown in Figure \ref{fig:isolet_improv} third panel). In many cases, the classifiers obtained their best results as $|F|$ decreased.\\

\begin{table}
	\begin{minipage}{0.45\linewidth}
		\resizebox{\textwidth}{!}{%
		\begin{tabular}[width=\linewidth]{@{}lcccc|rrrr@{}}
\toprule
 & \multicolumn{4}{c|}{\textbf{Naïve Bayes}} & \multicolumn{4}{c}{\textbf{SVM}} \\ \hline
 & \multicolumn{2}{c}{\textbf{AUROC}} & \multicolumn{2}{c|}{\textbf{Sensitivity}} & \multicolumn{2}{c}{\textbf{AUROC}} & \multicolumn{2}{c}{\textbf{Sensitivity}} \\ \hline
\textbf{|F|} & \multicolumn{1}{l}{\textbf{mean}} & \multicolumn{1}{l}{\textbf{std}} & \multicolumn{1}{l}{\textbf{mean}} & \multicolumn{1}{l|}{\textbf{std}} & \multicolumn{1}{l}{\textbf{mean}} & \multicolumn{1}{l}{\textbf{std}} & \multicolumn{1}{l}{\textbf{mean}} & \multicolumn{1}{l}{\textbf{std}} \\
\textbf{178} & 0.932 & 0.02 & 0.89 & 0.057 & 0.932 & 0.02 & 0.89 & 0.057 \\
\textbf{54} & 0.924 & 0.031 & 0.876 & 0.064 & 0.894 & 0.018 & 0.800 & 0.035 \\
\textbf{45} & 0.926 & 0.017 & 0.880 & 0.034 & 0.890 & 0.019 & 0.791 & 0.037 \\
\textbf{36} & 0.924 & 0.018 & 0.876 & 0.037 & 0.890 & 0.021 & 0.791 & 0.043 \\
\textbf{27} & 0.922 & 0.014 & 0.871 & 0.029 & 0.881 & 0.018 & 0.773 & 0.037 \\
\textbf{18} & 0.912 & 0.022 & 0.853 & 0.046 & 0.867 & 0.035 & 0.742 & 0.071 \\
\textbf{13} & 0.906 & 0.022 & 0.840 & 0.046 & 0.852 & 0.038 & 0.711 & 0.079 \\
\textbf{9} & 0.898 & 0.027 & 0.822 & 0.052 & 0.835 & 0.035 & 0.680 & 0.071 \\ \bottomrule
\end{tabular}}
\caption{Classification results for Epileptic Seizure Dataset with NB and SVM classifiers. Mean and Standard deviations are averaged over 5 trials.}
\label{tab:seizure}
	\end{minipage}\hfill
	\begin{minipage}{0.5\linewidth}
		\centering
		\includegraphics[width=\textwidth]{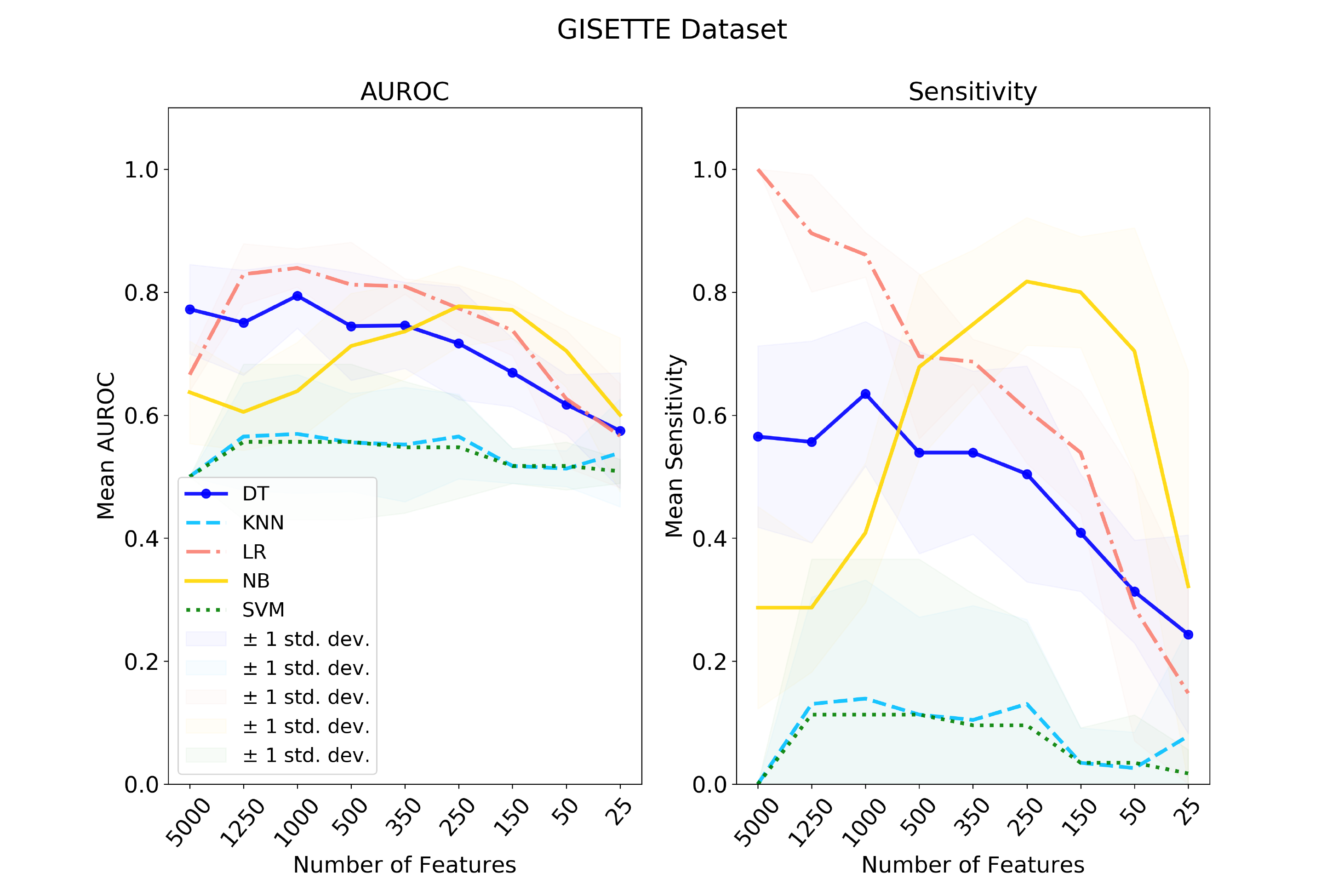}
		\captionof{figure}{Classification results for GISETTE Dataset with all 5 classifiers. Mean and Standard deviations are averaged over 5 trials.}
		\label{fig:gisette}
	\end{minipage}
\end{table}

In Table \ref{tab:seizure} we report the classification performance on the Epileptic Seizure Dataset. The first line summarizes baseline results. Note that we chose to include only NB and SVM classifiers, as LR, KNN and DT demonstrated a baseline performance that was too poor to meaningfully consider them for classification on this data. On the contrary, NB and SVM showed a high baseline performance despite the strong imbalance. Decreasing the amount of features used in classification did not hinder the performance, while reducing the dimensionality of the problem. For example, by reducing it to a third ($|F|=54$), NB did not significantly reduce AUROC or Sensitivity metrics, while the performance for much smaller subsets ($|F|=\{18,13,9\}$) remains comparable with the baseline.
In Figure \ref{fig:gisette} we report the results of the experiment on Gisette data. Note that this dataset was designed for feature selection benchmarking experiments, by including 2,500 predictive features and 2,500 probes. By looking at $|F|$ values on the x-axis, one can note that the feature subsets selected by the DSAEE are way smaller than the original number of noisy features. However, irrespectively of the baseline performance with $|F|=5,000$, all classifiers showed an increase in performance for some $|F|$ values. This could mean that the algorithm is first correctly excluding noisy features; then, among the informative predictors, it is progressively excluding correlated or reduntant features, identifying the most useful for the classification task at hand. This hypothesis is well supported by the behavior of NB classifier, that by design requires conditional independence to reach optimal classification \cite{zhang2005exploring}. In this experiment, NB yields a steep increase on both metrics for smaller $|F|$ values. Only LR suffered a steep decrease in Specificity, that was however balanced by the significant improvement in AUROC (meaning that the performance is better balanced between the two classes) for subsets between 1,000 and 250 features.

\subsection{Feature Selection Benchmarking Experiment} 
\label{benchmark}
We selected the benchmark feature selection methods for the performance comparison with our DSAEE approach s.t. they would be representative of different types of algorithms. In particular, we included (i) Chi-squared, a supervised filtering feature selection method based on univariate $\chi^{2}$ statistical tests, and (ii) Recursive Feature Elimination (RFE) \cite{guyon2002gene} - supervised wrapper method, that when combined with SVM classifier (RFE-SVM) was proven one of the best performing methods in \cite{maldonado2014feature} for feature selection in imbalanced settings. Finally, we also included (iii) Concrete AutoEncoder Feature Selector (CAEFS), an unsupervised feature selection method based on AEs\footnote{The number of features selected is defined as the \textit{K} nodes of the concrete layer. To compare the two models we trained the each DSAE learner and the CAEFS for the same number of epochs using the same batch size, and the architecture of the decoder was built equal to that of the DSAEs.}, that in \cite{concrete} was proven superior to most related algorithms mentioned in Section \ref{sec:aefs}.\\
\indent All benchmark methods were applied to the FSDS imposing a number of selected features equal to the features selected by DSAEE for the different $\delta$ levels, then the subsets of selected features were extracted from the CDS to test classification accuracy.
We compared the performance on Isolet dataset and Fashion MNIST dataset averaging on 5 trials for each experiment. In both cases we trained an ensemble of $B=25$ DSAEs.\\
\begin{figure}[t]
\centerline{\includegraphics[width=\columnwidth]{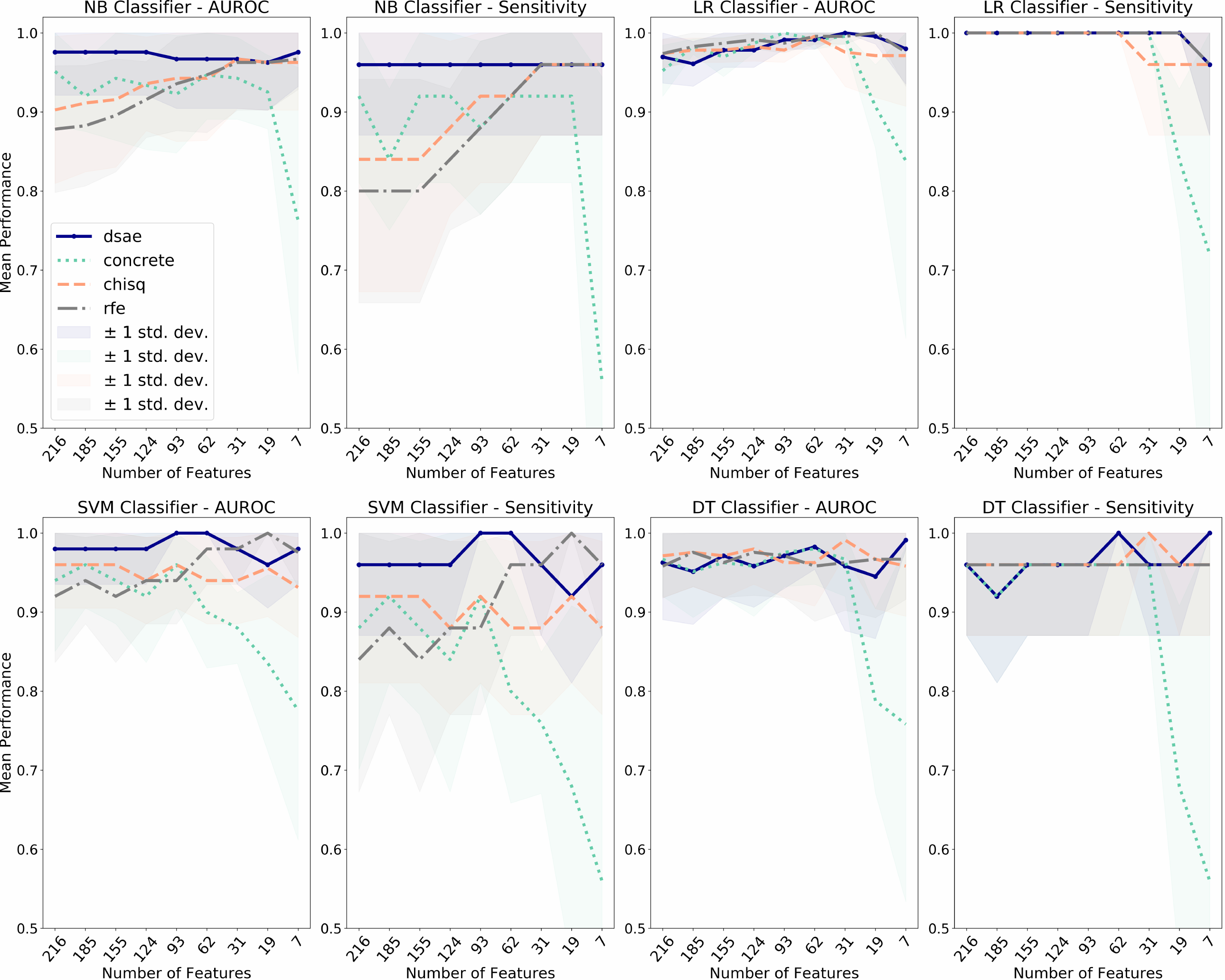}}
\caption{Classification benchmarking against other FS methods for ISOLET Datasets, for NB, LR, SVM and DT classifiers. Each classifier has one plot per metric (AUROC on the left, Sensitivity on the right)}
\label{fig:isolet_benchmark}
\end{figure}
In Figure \ref{fig:isolet_benchmark} we report the results on Isolet using four different classifiers, on Sensitivity and AUROC. 
Varying the threshold $\delta$ we selected a different subset of variables: the cardinality ($|F|$) of such subsets is reported on the x axes.
For what concerns NB and SVM classifiers, the DSAEE performed better than the competitors for almost all variables subsets on both indicators. In particular, it significantly outperformed the unsupervised CAEFS for smaller subsets, while the major competition on the smallest dimensionalities was represented by the supervised RFE. Note that RFE-SVM is a feature selection method proved among the best performers for imbalanced settings \cite{maldonado2014feature}, and the DSAEE either surpasses or reaches comparable performance levels in most cases  (see the two plots in the left bottom part of Figure \ref{fig:isolet_benchmark}). Similar results were obtained with the NB classifier. Regarding LR classifier, all methods seemed to perform well on this dataset, but our methodology reaches an almost perfect score on Sensitivity irrespectively of the threshold level, up until to only 7 variables, where the other AE-based FS method (CAEFS) lowered its average performance. These levels of Sensitivity and AUROC - irrespectively of the adopted classifier - on a dataset with significantly small sample size and extremely high dimensionality testify in favour of the applicability of our methodology in many real-life scenarios where the collection of observations might be costly or difficult.\\
\begin{figure}[t]
\begin{center}
\centerline{\includegraphics[width=0.65\columnwidth]{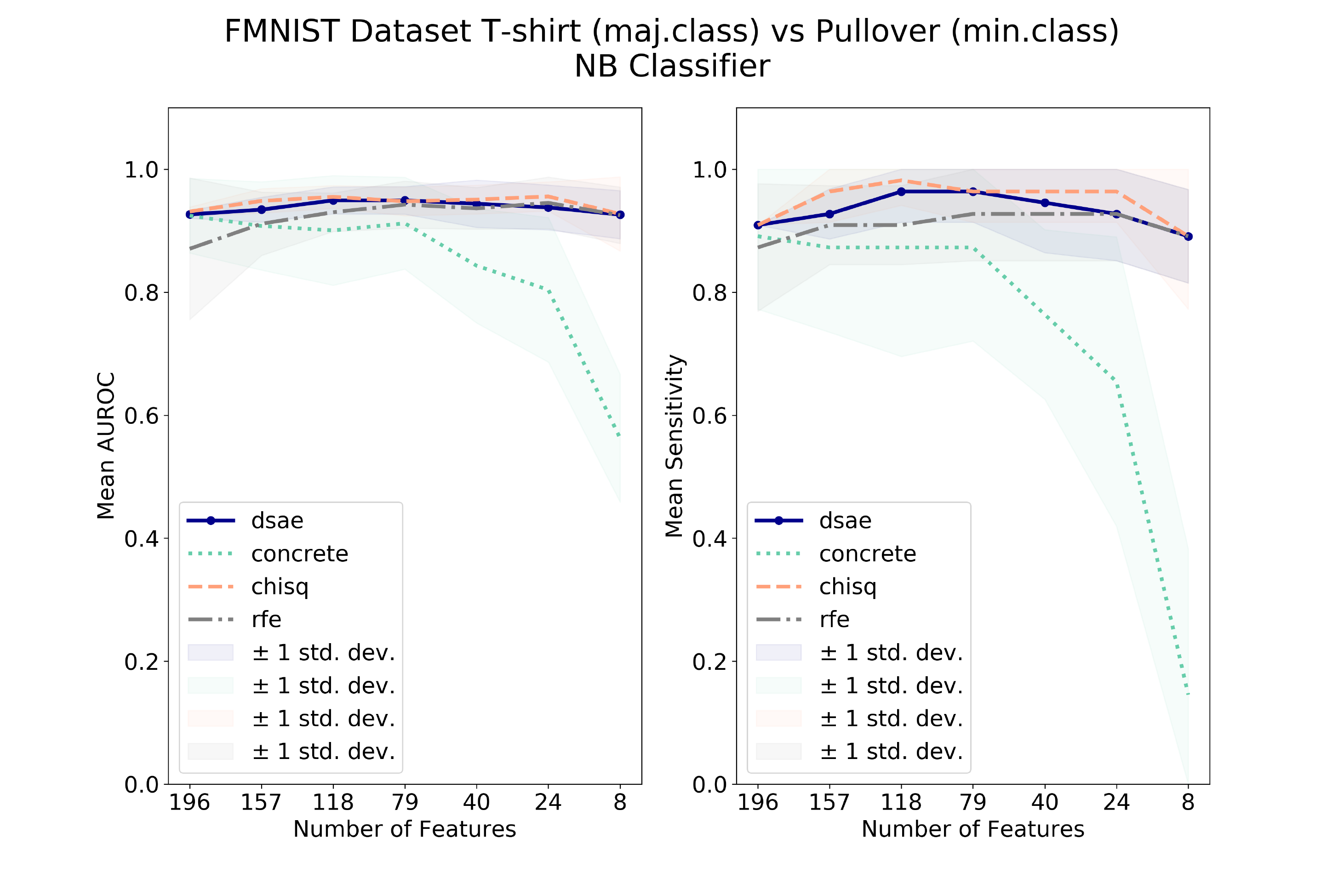}}
\caption{Classification benchmarking against other FS methods for Fashion MNIST dataset using NB classifier.}
\label{fig:fmnist_benchmark}
\end{center}
\end{figure}

\begin{table}[]
\centering
\resizebox{0.85\textwidth}{!}{%
\begin{tabular}{@{}llccccc@{}}
\toprule
Dataset &  & \textbf{DSAE} & \textbf{CHISQ} & \textbf{RFE} & \textbf{CONCRETE} & \textbf{DSAE PARALL.} \\ \midrule
{\textbf{ISOLET}} & Average Time {[}min{]} & 14.010 & 0.046 & 1.246 & 117.268 & 0.560 \\
 & Std {[}min{]} & 5.927 & 0.005 & 0.048 & 78.034 & 0.237 \\ \midrule
{\textbf{FMNIST}} & Average Time {[}min{]} & 7.687 & 0.087 & 22.852 & 0.618 & 0.256 \\
 & Std {[}min{]} & 0.201 & 0.001 & 1.424 & 0.017 & 0.007 \\ \bottomrule
\end{tabular}%
}
\caption{Comparison of average runtime performance of all benchmark methods on the Isolet Dataset. The average time is computed considering total process time to select feature subsets for all $\delta$ thresholds, and averaged over 5 trials. }
\label{tab:time}
\end{table}

In Figure \ref{fig:fmnist_benchmark} we compare the performance of the DSAEE on FMNIST dataset using the best performing classifier in terms of performance improvement (Figure \ref{fig:isolet_improv}). Our algorithm confirmed its superiority w.r.t. the competing AE-based FS method, while keeping a comparable performance to the other benchmark algorithms, all set to a very high performance up until an extremely small feature subset (8 pixels from the original 784).\\
Note that, as can be noticed from Figure \ref{fig:isolet_improv}, both datasets allowed for high prediction accuracy on both classes even before feature selection. This indicates that probably, despite the imbalanced setting, the two classes are sufficiently separated and consistently characterized to allow classifier to correctly separate them and generalize well just by seeing few examples of the underrepresented class. For this reason, it is not surprising to see all algorithms (especially the supervised ones) perform quite well on this feature selection and classification task. Nonetheless, although our ensemble algorithm is based on unsupervised learners, it consistently reached or surpassed the supervised approaches, and performed significantly better than the unsupervised one.\\
In Table \ref{tab:time} we report the total process runtime to complete all feature subsets selections (for all $\delta$ values) for the different algorithms, averaged over all trials. In the first column each of the trained DSAEs are processed in sequence, while in the last one we report the estimated average time to perform the algorithm's training in parallel. Even though the sequential training time is not prohibitive per se, its parallelized version outperforms the wrapper RFE and the other AE-based algorithm (CAEFS) by far, while enjoying the beneficial robustness of an ensemble framework. 

\subsection{Interpretability}
\begin{figure}[t]
\begin{center}
\centerline{\includegraphics[width=\columnwidth]{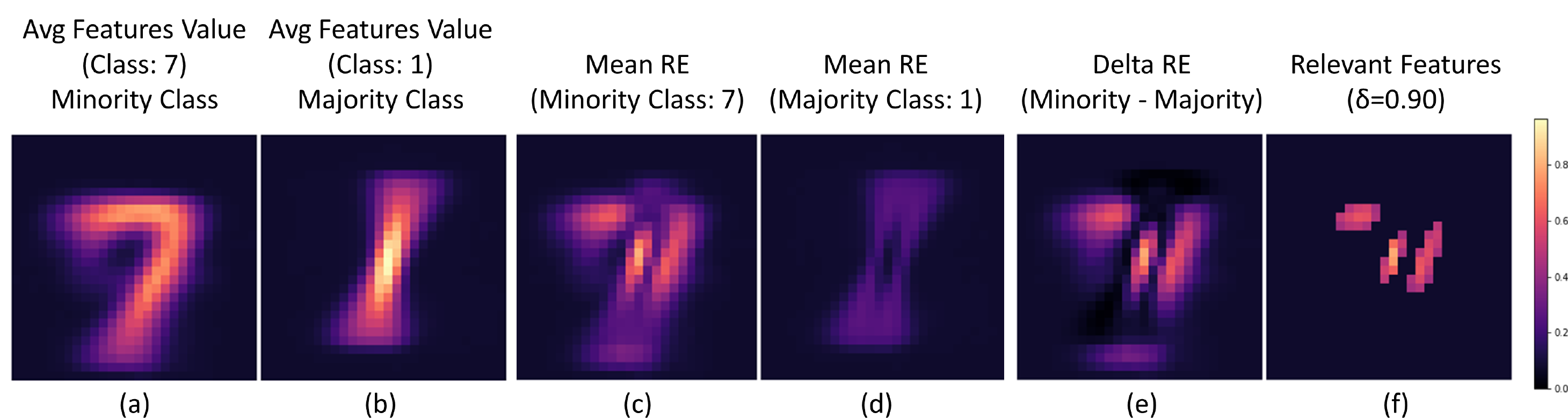}}
\caption{Results of the experiment on the 7 (minority) and 1 (majority) classes. In these 28x28 pixels images each pixel represents a feature. Subfigures (a) and (b) represent the mean of all values the two classes take in the FS dataset. The color scale is shared across all six subfigures. Subfigure (c) reports the average RE for the minority class, while (d) is the representation of the majority class average RE - Note that being the '1's class the majority one, the model learns to reconstruct precisely the center of the vertical line that draws the digit. The vector $\Delta$ is reported in (e), while (f) depicts the selected variables with a threshold $\delta = 0.9$.}
\label{fig:17experiment}
\end{center}
\vskip -0.2in
\end{figure}

\begin{figure}[t]
\centerline{\includegraphics[width=\columnwidth]{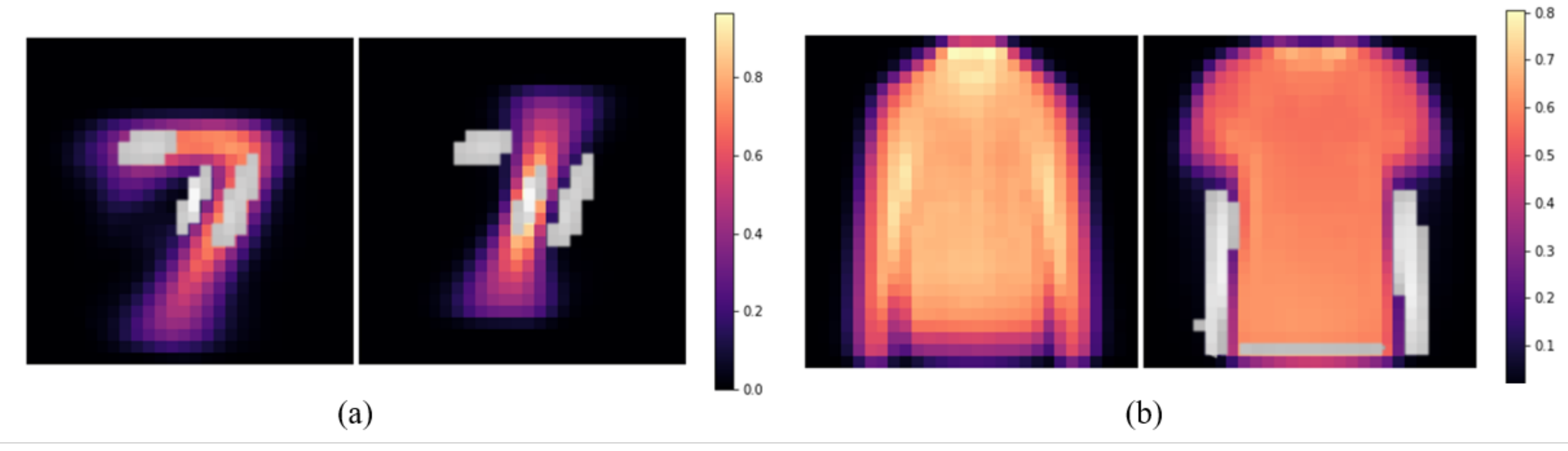}}
\caption{(a) MNIST Dataset. The most relevant identified features are represented in gray scale over the average minority (right) and majority class (left) representations. (b) Fashion MNIST Dataset. Here, it is clear how the most relevant features to distinguish \textit{coats} from \textit{t-shirts} are the pixels that compose the sleeves of the coat.}
\label{fig:overlapped}
\end{figure}

One advantage of FS for classification lies in the increased interpretability of the subsequent algorithms and results. Indeed, identifying features that are the most informative, w.r.t. a target class within a dataset is an insightful information by itself in many application contexts. In the era of black-box classifiers, a reduction in the amount of information fed to these algorithms is per se a way of improving the interpretability of (and the control over) the obtained classifications. In the case of our proposed algorithm, the selected features are the subset of variables where the minority class distances the majority one the most.\\
In Figure \ref{fig:17experiment} we report some visualizations from the MNIST Dataset that help in understanding the feature selection process performed by our algorithm. The small set of selected features for $\delta=0.90$ (Figure \ref{fig:17experiment}.f) is then overlapped (in gray scale) to the average representation of the two classes (Figure \ref{fig:overlapped}.a). This visualization allows us to recognize how the selected features include all pixels where the minority class ('7' digits) have different characteristics w.r.t. the '1' digits class.\\
In Figure \ref{fig:overlapped}.b we propose the same visualization for the Fashion MNIST dataset. Note that these features subsets were obtained in an highly imbalanced setting, as reported in Table \ref{tab:datasets_proportions}, but the selected features are extremely meaningful nonetheless.
\label{interpretation}

\subsection{Case Study application in Radiogenomics}
\label{radiogenomics}
Class imbalance is a daunting issue in many real life applications, especially when dealing with medical and biological data \cite{thabtah2020data} (cfr. Section \ref{intro}). So far, we presented simulation studies and proofs of concept to demonstrate the generalizable potential of the proposed algorithm. However, the value of the presented approach lies in its demonstrated applicability to complex scenarios arising from real life research settings. 
Indeed, in this section we present a real data application of the DSAEE FS algorithm in the field of radiogenomics. A detailed report on the study can be found in \cite{massi2020deep}. However, because of the aforementioned reasons, we were interested in providing here a brief description nonetheless. Specifically, we focused on the long term outcomes of radiotherapy on patients suffering from prostate cancer. The final aim was to validate genetic locations (in the form of Single Nucleotide Polymorphisms, or SNPs) that can be associated with Late Toxicity (LT) outcomes. Experts were indeed interested in finding whether among the features (i.e. the SNPs) with high association to the 5 considered LT endpoints in previous studies on different cohorts, some could be validated as relevant for the cohort at hand ($\sim$ 1,700 patients with an incidence of the positive class always below $10\%$ for each endpoint and a total number of 43 SNPs to evaluate). We applied our DSAEE on each of the 5 endpoints separately, and we selected SNPs with different $\delta$ thresholds ($\delta = \{0.7, 0.8, 0.9, 0.95\}$). This being an unsupervised setting it is hard to comment on precision of the results without the required clinical expertise. However, notably, for one of the endpoints (i.e. Late Urinary Frequency) 3 SNPs identified as relevant by our method for all $\delta$ values were previously mentioned in literature \cite{kerns2016meta} as the most strongly associated to this endpoint.\\This is an interesting application case in which FS methods are useful to profile minority class, and provide useful insights to researchers. As introduced in Section \ref{intro}, our FS algorithm is indeed tailored to respond to similar needs and to deal with complex scenarios where the class of interest is extremely rare. 

\section{Discussion and Conclusions}
\label{conclusion}
In this paper we presented a Deep Learning-based ensemble approach to select features for highly imbalanced classification tasks. The proposed approach exploits Deep Sparse AutoEncoders as \textit{weak learners}, each trained to learn the \textit{normal} patterns in majority class observations, and tested on both majority and minority class data. Diversity among components of the ensemble is fostered by a tailored sampling procedure and the sparsity constraint on the training loss function. Features are ranked averaging on the RE of the ensemble of learners to identify the most informative ones, where minority class distribution differs from majority class the most.\\
We performed a series of experiments to test the potential of our DSAEE. First, we verified the capability of our algorithm to avoid the degradation of classification performance induced by selecting feature subsets in a setting of strong imbalance \cite{yin2013feature}. We compared baseline performances with that obtained with subsets of increasingly small dimensionality, using a wide range of datasets with different characteristics to simulate diverse research application scenarios. Then, we benchmarked our method against other feature selection methods, demonstrating the superior or comparable performance of the DSAEE feature selector. Note that most of the algorithms we compared the DSAEE with had the advantage of being supervised, or even tailored to maximize prediction accuracy on minority class (RFE-SVM).
\\
Our FS algorithm is tailored to manage extremely imbalanced settings with the aim of attaining all the advantages of FS methods without sacrificing too much on the classification performance by reducing the amount of information supplied to classifiers. In some cases, the algorithm was capable of identifying subsets of the original features yielding an improved performance in terms of AUROC and/or Specificity (cfr. Figure \ref{fig:isolet_improv} and Figure \ref{fig:gisette}). In particular, the improved Specificity might be induced by the training procedure of each ensemble DSAE component: indeed, AEs by nature represent an approximation of the identity function and the applied model is compelled to learn the common characteristics of the data \cite{sarvari2019unsupervised}. By training on majority class only, the learnt data distribution does not include the characterization aspects of minority class instances, thus generating higher reconstruction errors on those features. Moreover, the initial data sampling, once included in an ensemble framework, allows to extract reliable information even when the observations belonging to the minority class are limited. While creating the different sampled training and test set for each ensemble component, the minority class is indeed studied against various subsets of the majority one, thus enhancing the informative power of the small underrepresented sample.\\
On top of the sampling procedure we included to the training loss function of our components a sparsity penalty term, that besides fostering components' diversity reduces the need for lengthy optimization of the DSAEs' architecture. Indeed, the penalty term forces the number of active nodes in the hidden layer to adapt to the sample of training data, reducing autonomously the risk of learning trivial representations.\\
Besides all the above, the DSAEE Feature Selection algorithm is a filtering method, meaning that it is agnostic to the classifier exploited to discriminate between classes. This may slightly hinder classification accuracy compared for instance to wrapper methods, but gains generalizability of the identified features. Moreover, when compared to wrapper methods, our approach does not incur in the risk of sub-optimal solutions in high-dimensional settings, where evaluating all possible combinations of features would be computationally intractable.
When compared to embedded methods, our AE-based approach is capable of capturing nonlinear relationships among features. Kernel-based embedded feature selection methods were proposed to learn nonlinear representations \cite{liang2006feature}, but they are limited by the fixed kernel, and the choice of the optimal kernel or combination of kernels is not straightforward.\\

In conclusion, with this work we are taking inspiration from different methodological domains to develop a novel filtering feature selection algorithm that is (i) robust thanks to its ensemble nature, (ii) capable to learn complex patterns in data because of its AE components, (iii) provides interpretable insights and (iv) is specifically tailored to tackle class imbalance. All these considerations promote the usefulness of our DSAEE feature selector in real-life contexts where data are imbalanced, minority class observations have great relevance, sample size is small, and interpretability of results is crucial. We provided a direct example in Section \ref{radiogenomics}, where a real data application is briefly described.\\
Future works might be devoted to studying the applicability of the DSAE feature selector to imbalanced multi-class classification problems or to further develop the analysis of the RE distributions to select features.


\section*{Acknowledgments}
The Authors thank the ERA PerMed Cofund program, grant agreement No ERAPERMED2018-44, RADprecise - Personalized radiotherapy: incorporating cellular response to irradiation in personalized treatment planning to minimize radiation toxicity.

\subsection*{Financial disclosure}
FG was funded by the UK Medical Research Council programme\\ MRC$\_$MC$\_$UU$\_$00002/5.

\subsection*{Conflict of interest}
The authors declare no potential conflict of interests.

\appendix
\section{DSAE Architectural and Implementation Details\label{Appendix1}}
In the following section we provide the details of the architectural and implementation choices made on the DSAEs for the different experiments. Note that these choices are provided for the sake of results' reproducibility, but they should not be considered a strict guideline about how the components in the ensemble should be built. Indeed, the DSAE is a fundamental building block of our methodology, but just as in any application of deep learning models, it should be customized to the problem at hand. For that reason, we did not focus all our effort in seeking for the lightest and fastest possible architecture. Our focus and concern was on the demonstration of the potentials of the methodology as a whole, that exploits this well known building block within a novel algorithm to robustly identify features to separate the two classes.\\\\
The algorithm was developed in Python 3.6, using Keras and Tensowrflow as back-end. The code was run on Jupyter notebooks hosted on Google Colab Virtual Machines\footnote{\url{https://colab.research.google.com/notebooks/intro.ipynb}}, with access to GPUs and Fast VMs thanks to the Pro subscription. The types of GPUs that are available in Colab vary over time. The GPUs available in Colab often include Nvidia K80s, T4s, P4s and P100s. There is no way to choose what type of GPU you can connect to in Colab at any given time.
\\\\ 
An overview of the architectural and implementation choice of the proposed method for the four different datasets evaluated in this work is reported in Tab.\ref{app:tab:architecturalchoice}. In this table we give details about the Encoder (number of nodes for input and hidden layers, activation function per layer) and the Decoder (number of nodes in the hidden and output layers, activation function per layer). If a single type of function is reported (like tanh as activation function in the Decoder for ISOLET data), this means that we chose the same activation function across all layers. In the bottom part of Tab.~\ref{app:tab:architecturalchoice}, we describe the number of epochs, the batch size and the parameter B of the algorithm. 

The following parameters have been selected consistently across the four different datasets:
\begin{itemize}
    \item the last hidden layer in the Encoder had an $L_{1}$ penalization on the activation of the 200 nodes, with $\lambda = 10e^{-5}$. The value of this hyperparameter was chosen between $\lambda = (10e^{-5}, 10e^{-10}, 10e^{-20})$ as the one that guaranteed a low reconstruction error, while favouring a sufficient penalization on the activation of the hidden nodes. 
    \item the DSAE was trained with the Adam optimization algorithm ($learning\_rate = 0.001$, $\beta_{1}=0.9$, $\beta_{2}=0.999$). We decided not to optimize these hyperparameters because of computational time of the experiments. Therefore we kept them as the default standard suggested by literature when analyzing all the four different datasets.
\end{itemize}

\begin{sidewaystable}
\caption{Details of the architectural and of the implementation are reported here for the four analysed datasets. The function tanh is the hyperbolic tangent, the ReLu is the Rectified Linear Unit function. \textit{Enc.} section of the table reports the encoder architecture, while \textit{Dec.} details the decoder. The bottom part of the table (\textit{Train.}) reports details on the training procedure (number of epochs, batch size and number of ensemble components $B$).}
\centering
\begin{tabular}{|c|c|c|c|c|c|c|} 
\hline
\multicolumn{2}{|l|}{}                                                                    & \textbf{ISOLET} & \textbf{GISETTE} & \textbf{MNIST} & \textbf{F-MNIST}  & \textbf{EP. SEIZURE} \\ 
\hline
\multirow{3}{*}{Enc.}  & Nodes & 617-600-500-250-200 & 5000-1000-500-250-250 & 784-700-500-250-200 & 784-700-500-250-200 &  178-132-64-32       \\ 
\cline{2-7}
                          & \begin{tabular}[c]{@{}l@{}}Act.\\funct.\end{tabular} &   tanh-tanh-tanh-ReLu & sigmoid-sigmoid-ReLu-ReLu & sigmoid & tanh-tanh-tanh-ReLu  & tanh       \\ 
\hline
\multirow{3}{*}{Dec.}  & Nodes  &
250-500-600-617     &
250-500-1000-5000  &
250-500-700-784     &
250-500-700-784      & 64-132-178 \\ 
\cline{2-7}
                          & \begin{tabular}[c]{@{}l@{}}Act.\\funct.\end{tabular} &  tanh    & ReLu-sigmoid-sigmoid-sigmoid & sigmoid       &   tanh    & tanh   \\ 
\hline
\multirow{3}{*}{Train.} & Epochs                                                        &  100 & 50 & 50 & 100 & 200  \\ 
\cline{2-7}
                          & Batch sz.  &  10 & 1000 & 100 & 100 & 1000 \\ 
\cline{2-7}
                          & B & 25 & 25 & 50 & 50 & 30       \\
\hline
\end{tabular}%
\label{app:tab:architecturalchoice}
\end{sidewaystable}

\newpage

\bibliography{biblio}%

\end{document}